\documentclass[journal]{IEEEtran}
\usepackage{amsmath,amsfonts}
\usepackage{algorithm}
\usepackage{array}
\usepackage[caption=false,font=normalsize,labelfont=sf,textfont=sf]{subfig}
\usepackage{textcomp}
\usepackage{stfloats}
\usepackage{url}
\usepackage{verbatim}
\usepackage{graphicx}
\usepackage{cite}
\usepackage{booktabs}
\usepackage{tabularx}
\usepackage{array}
\usepackage{multirow}
\usepackage{amsmath}
\usepackage{graphicx}   
\usepackage{subcaption} 
\usepackage{makecell}
\usepackage{placeins}
\usepackage{afterpage}
\usepackage{algorithm}
\usepackage{algpseudocode}
\usepackage{siunitx}
\usepackage{amssymb} 
\usepackage{amsmath, amssymb, amsthm}

\usepackage{amsthm} 

\begin{document}

\title{Dual-Domain Manifold Modeling for Hyperspectral Image Fusion}

\author{Chengxin Xie, Qiya Song,  Yangbangyan Jiang, Renwei Dian, Xudong Kang
\thanks{This work was supported in part by the National Natural Science Foundation of China under Grant 62401204. 
\par
Chengxin Xie and Qiya Song are with the College of Information Science and Engineering, Hunan Normal University, Changsha, China (e-mail: chengxin@hunnu.edu.cn; sqyunb@hnu.edu.cn). 
\par
Yangbangyan Jiang is with Institute of Information Engineering, Chinese Academy of Sciences,Beijing, 100000, Beijing, China. \par 
Renwei Dian and Xudong Kang are with the School of Artificial Intelligence and Robotics, Hunan University, Changsha, China. }
}

\maketitle

\begin{abstract}
Achieving a coherent integration of spectral richness and spatial fidelity remains a central objective in hyperspectral image fusion. However, existing hyperspectral image fusion methods struggle to effectively model geometric constraints. In the spatial domain, weak spatial–spectral interaction limits geometry-aware feature learning and suppresses high-frequency structural information, resulting in low-frequency bias and structural degradation. In the spectral domain, local manifold structures induced by spectral similarity are insufficiently exploited, limiting intrinsic pixel relationship modeling and fine-grained spectral reconstruction. To address these challenges, we propose a dual-domain manifold modeling (DDMM) framework. Specifically, we introduce a Topology-Aware Transformer (TPFormer) that combines global attention with neighborhood propagation, jointly modeling spatial topology and pixel-level feature manifold relationships to capture intrinsic spatial–spectral structures and improve topology-aware representation learning. Furthermore, a Frequency-Decoupled Spatial–Spectral Collaborative Fusion (FDSCF) module is devised, in which features are projected into the frequency domain via the discrete cosine transform and explicitly decoupled into low- and high-frequency components. Guided by a low-rank structural prior and spectral-driven spatial enhancement, FDSCF selectively enhances geometry-aware high-frequency features, strengthening spatial–spectral coupling and recovering sharper edges and finer textures.
Extensive experiments on multiple benchmark datasets demonstrate that DDMM achieves superior overall performance over state-of-the-art methods in terms of spatial structure preservation and spectral reconstruction. 
\end{abstract}


\section{Introduction}
\IEEEPARstart{H}{yperspectral} imaging aims to construct high-dimensional spatial-spectral representations of ground objects, yet the physical limitations of sensors prevent a single modality from achieving both high spatial and spectral resolution \cite{liu2025selective,9992028}. Information fusion has therefore become a prevailing reconstruction paradigm, where auxiliary modalities compensate for missing information in the primary modality, enabling cross-modal enhancement \cite{chen2025hyperspectral,yan2025hyperspectral}. As a representative multi-source reconstruction approach, hyperspectral image fusion leverages the complementarity between hyperspectral and multispectral data to produce images with both rich spectral fidelity and fine spatial detail, and has emerged as a central technique in computational imaging \cite{liu2025selective}. The demand for hyperspectral image fusion across various application domains is growing significantly \cite{11597948}. These applications range from remote sensing monitoring systems that require improved land-cover recognition accuracy, to environmental perception platforms dependent on the synergistic analysis of multi-source data, and further to urban planning support systems demanding efficient integration of complementary information \cite{11261642}. All involve complex couplings of spectral and spatial features, framed as high-dimensional feature fusion problems \cite{liu2025low,9992028}. However, preserving spectral fidelity and spatial structural coherence during fusion, while modeling geometric features and nonlinear mapping relationships, remains a significant challenge in remote sensing data processing \cite{10681101,10874856,chen2025hyperspectral,11071942}.\par
Hyperspectral image fusion methods are generally classified into three categories: multi-resolution analysis-based, prior modeling-based, and deep learning-based approaches \cite{10703123,9874905}. Multi-resolution analysis approaches aim to generate hierarchical image representations in the frequency or spatial domain via multi-scale decomposition, enabling beneficial spatial detail injection while preserving spectral information \cite{yu2025transformer,yan2025hyperspectral}. However, these methods often introduce edge blurring or structural distortions when processing non-stationary textured regions and struggle to adapt to geometric deformations in complex scenes \cite{yu2025transformer}. In contrast, prior modeling-based methods use prior knowledge to model the image degradation process, thereby improving the physical consistency and robustness of reconstruction results \cite{10198475,11247869}. Recently, deep learning-based methods, leveraging their strong nonlinear fitting capabilities, have learned cross-modal feature mappings in an end-to-end manner, significantly improving fusion accuracy and generalization performance \cite{10681101,10188591}. Nevertheless, both conventional deep learning-based models and prior-based methods often fail to maintain structural integrity in complex terrestrial scenes \cite{10275100,9641801,chen2022decomposition}. These limitations hinder their ability to generalize in high-dimensional feature spaces \cite{10681101}. \par
In real-world settings, hyperspectral images often exhibit complex geometric constraints and nonlinear manifold structures, where spatial deformations at object boundaries reflect pixel-level correspondences across modalities, and fine-grained spectral variations convey rich material information \cite{yu2025transformer,yan2025hyperspectral,10443302}.
To capture these intrinsic features, prior studies have employed attention mechanisms to enhance cross-modal associations in local regions, or leveraged physical constraints to guide image reconstruction, thereby improving the reliability of the results \cite{10035506,11276863}. Recently, Transformer-based architectures have been widely adopted to enhance long-range spatial–spectral interactions through self-attention mechanisms \cite{11277353}. However, self-attention primarily establishes data-dependent feature correlations based on global affinity, while lacking explicit constraints regarding the underlying geometric structure of hyperspectral manifolds. Consequently, highly similar but spatially or spectrally inconsistent pixels may participate in information exchange, which potentially weakens local manifold continuity and introduces structural inconsistency during reconstruction. Beyond the limitations of attention-based feature interaction, existing methods still struggle to effectively model geometric constraints. Spatially, they frequently suffer from \textit{inconsistent edge preservation and distorted structural boundaries}, which undermine the fidelity of spatial details and lead to blurred or misaligned features in complex scenes. Spectrally, the \textit{insufficient exploitation of pixel-level manifold structures} derived from spectral similarity limits the recovery of fine spectral signatures, restricting their ability to dynamically adapt to geometric deformations and spectral variations across diverse scenarios \cite{yan2025hyperspectral}. 

\par
To address the above limitations, we propose a dual-domain manifold modeling (DDMM) framework, which jointly explores spatial topology and pixel-level feature manifold relationships to enhance spatial–spectral representation learning. Specifically, DDMM consists of two complementary components. First, the Topology-Aware Transformer (TPFormer) integrates global self-attention with neighborhood propagation to establish topology-guided feature interactions, enabling effective modeling of spatial topology and pixel-level feature manifold relationships for capturing intrinsic spatial–spectral structures. Second, the Frequency-Decoupled Spatial–Spectral Collaborative Fusion (FDSCF) module performs frequency-aware feature decomposition via the discrete cosine transform and selectively enhances geometry-aware high-frequency components under a low-rank structural prior, thereby alleviating low-frequency bias and improving structural detail preservation.
The main contributions are summarized as follows: \par
\begin{itemize} 
\item We propose a dual-domain manifold modeling (DDMM) framework for hyperspectral image fusion, which jointly models spatial topology and pixel-level feature manifold relationships to enhance spatial–spectral representation learning and improve reconstruction fidelity.
\item A Topology-Aware Transformer is developed that integrates global attention with neighborhood propagation to model spatial topology and pixel-level feature manifold relationships, capturing spatial–spectral structures and improving topology-aware representation learning.
\item  We design a Frequency-Decoupled Spatial–Spectral Collaborative Fusion (FDSCF) module that performs DCT-based frequency decomposition and selectively enhances high-frequency structural components guided by structural priors, improving spatial detail preservation.
\item Extensive experiments on three benchmark datasets demonstrate that the proposed method achieves superior overall performance over recent state-of-the-art approaches, validating its effectiveness in hyperspectral image fusion.
\end{itemize} 
\section{Related Work}
\subsection{Graph Convolutional Network-Based Methods}
Graph convolution networks (GCNs) have emerged as a powerful tool for multimodal fusion, enabling hierarchical representation learning of manifold-structured data by constructing structural graphs from hyperspectral and multispectral images. To address limitations in modeling spatial-spectral joint dependencies inherent in conventional CNN-Transformer architectures, Yang et al. \cite{yang2023gtfn} proposed the GCN-Transformer Fusion Network (GTFN), which leverages long-range spectral correlations through graph-based modeling. Nevertheless, GTFN exhibits limited capacity in capturing fine-grained spatial patterns and local textural details, thereby constraining its effectiveness in complex classification tasks. A majority of existing fusion strategies rely on basic upsampling operations, which often introduce reconstruction artifacts and induce spectral distortion due to inadequate geometric preservation. To mitigate these issues, Chi et al. \cite{11014272} introduced the Cross-Modal Spectral Graph Attention and Vision Transformer (CSGAV), a fusion framework that integrates graph attention mechanisms with Transformers to better maintain spectral integrity. However, CSGAV lacks explicit regularization for intermodal geometric consistency, frequently leading to spatial misalignment in the fused outputs. To enhance cross-modal discrepancy modeling, Katiyar et al. \cite{10715660} developed a Hierarchical Multilevel GCN-CNN Fusion (HMGCF) method, designed to simultaneously capture local spatial features and global contextual structures. Despite its representational power, HMGCF does not enforce constraints on cross-modal semantic alignment, making it prone to boundary blurring and limiting its adaptability to geometric deformations in heterogeneous scenes.
\subsection{Traditional model-Based Methods}
Traditional model-based fusion methods mainly exploit explicit priors, including low-rank, sparsity, and physical imaging constraints, to regularize hyperspectral image fusion. Liu et al. \cite{liu2025low} proposed the Low-Rank Transformer Network (LRTN), which jointly exploits spatial and spectral low-rank priors for fusion. To enhance the flexibility of low-rank modeling across modalities, Dian et al. \cite{10522984} introduced a generalized tensor nuclear norm that reduces the sensitivity to tensor mode permutations while improving low-rank representation capability. Yuan et al. \cite{10975051} developed an unsupervised feature fusion-guided network that integrates multi-scale semantic features and sparsity constraints to improve robustness against noise and outliers. Furthermore, several studies have incorporated physical image formation models into network architectures to enhance reconstruction reliability. Yu et al. \cite{yu2025otpnet} formulated fusion as an optimization process with deep priors and designed proximal splitting operators to exploit intrinsic spatial–spectral characteristics. Meanwhile, Huang et al. \cite{huang2026lrdun} introduced low-rank decomposition into a deep unfolding framework to alleviate the ill-posedness of hyperspectral reconstruction while improving efficiency. Despite these advances, existing methods still face challenges in preserving fine structural details in complex terrestrial scenes with significant spatial heterogeneity and material variations.

 \begin{figure*}[htbp]    
\centering     
\includegraphics[width=0.92\textwidth]{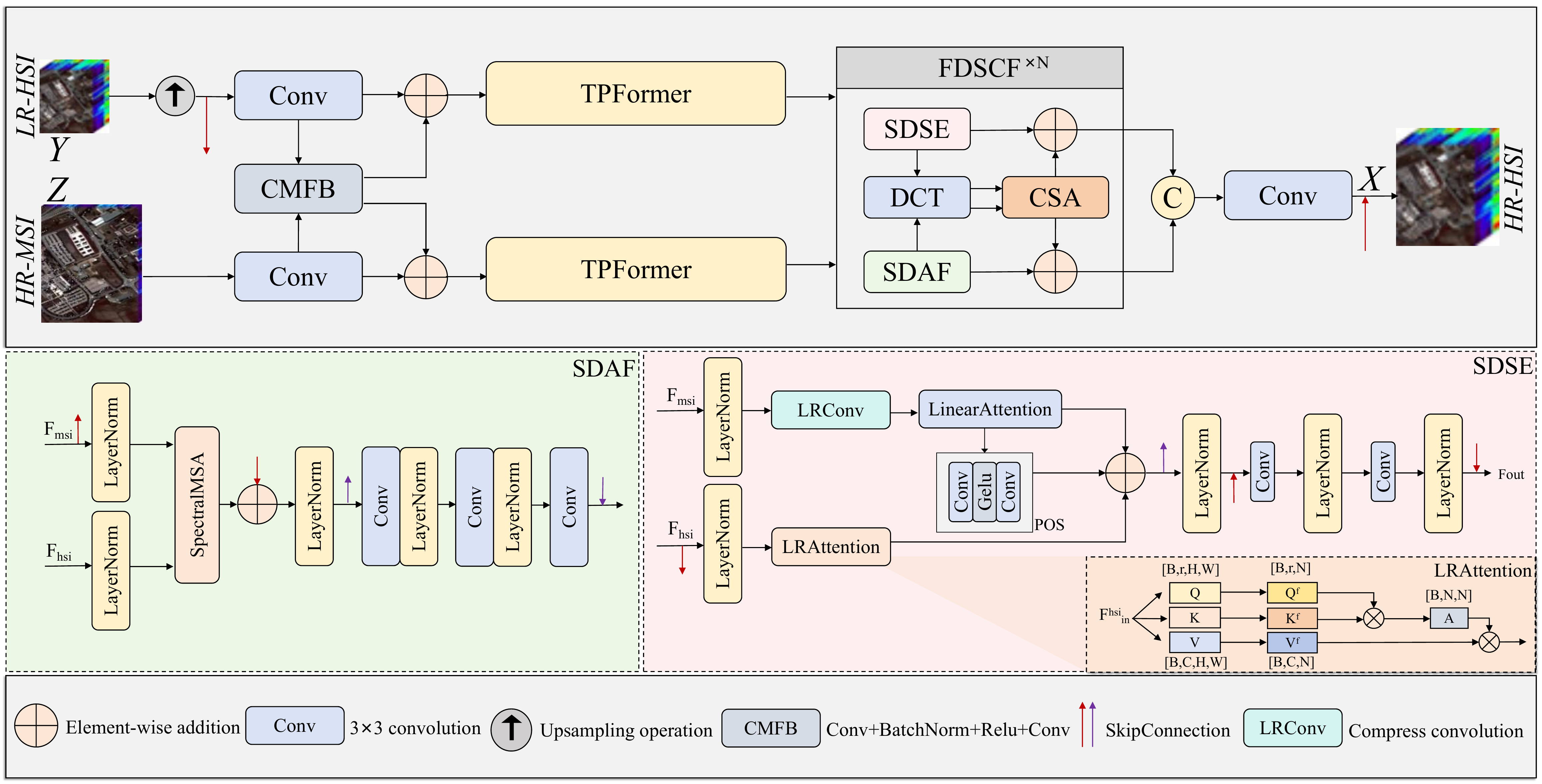} 
\caption{The illustrative workflow of the proposed Dual-Domain Manifold Modeling (DDMM) method. TPFormer jointly models spatial topology and pixel-level feature manifold relationships to enable fine-grained spatial–spectral representation learning. The FDSCF module is a dual-domain collaborative fusion mechanism that enhances high-frequency geometric details, thereby recovering sharp edges and fine textures.}
\end{figure*}

\subsection{Deep Learning-Based Methods}
Deep learning-based fusion methodologies encompass architectures such as convolutional neural networks, attention mechanisms, and Transformers, which perform fusion operations at pixel, feature, and decision levels \cite{wan2025remote,zhang2025adaptive,liu2022siamhyper}. To address the loss of high-frequency components in implicit neural representations, Liang et al. \cite{liang2024fourier} proposed a Fourier-enhanced implicit neural fusion network, which leverages spectral domain modeling to improve high-frequency signal representation and enlarge the receptive field. Yan et al. \cite{yan2025spatial} introduced a mutually guided spatial–spectral fusion network to capture the interactions between spatial and spectral domains, thereby enhancing fusion fidelity. However, under complex spatial–spectral degradation, existing methods still encounter spectral distortions and spatial artifacts. To alleviate these issues, Zhu et al. \cite{10555311} developed a self-supervised hyperspectral–multispectral fusion framework with a multi-scale fusion strategy for recovering fine-grained structural details. Meanwhile, several studies have explored adversarial learning paradigms to further improve perceptual quality. For example, Yu et al. \cite{yu2024unmixing} proposed an unmixing-prior-guided fusion method that utilizes spectral unmixing to simplify latent space modeling and incorporates abundance maps from unregistered images into generative networks to preserve spatial distribution characteristics. Beyond generative approaches, Liu et al. \cite{11347025} developed SEMFNet by integrating spectral–spatial edge enhancement with a partition-based Transformer for detailed reconstruction. However, although hyperspectral images exhibit complex geometric structures and nonlinear low-dimensional manifolds, existing deep learning methods often fail to explicitly model pixel-level feature manifold relationships. Consequently, their feature representations remain insufficiently adaptive to geometric deformations and spectral variations across diverse remote sensing scenarios.

\section{Proposed Method}
As shown in Fig. 1, the DDMM model comprises two core components:  the Topology-Aware Transformer (TPFormer), and the Frequency-Decoupled Spatial-Spectral Collaborative Fusion (FDSCF) module. The TPFormer module extracts global dependencies and fine-grained manifold features, which are integrated into the FDSCF module to establish a guided feature interaction mechanism. This process enhances the complementarity of cross-modal information. 
\par
\begin{figure*}[htbp]    
\centering  
\includegraphics[width=0.9\textwidth]{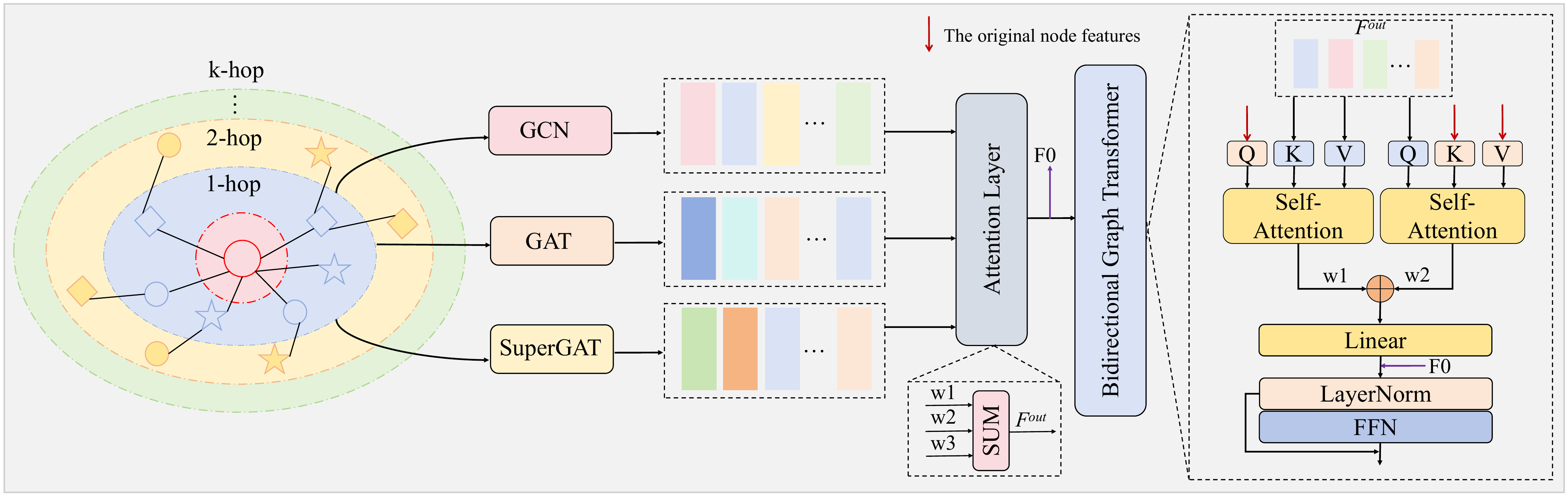} 
\caption{The block diagram of the Topology-Aware Transformer (TPFormer). Each pixel is treated as a graph node, whose attributes are represented by the corresponding spectral-aware feature vector extracted from the input feature map. \textbf{The original node features} indicates the feature matrix formed by all nodes in the graph.}
\label{fig:tpformer}
\end{figure*}
\subsection{Topology-Aware Transformer}
In hyperspectral and multispectral image fusion, the target HR-HSI can be interpreted as jointly generated from a LR-HSI through spatial degradation and a HR-MSI through spectrally selective responses. Let the ground-truth be denoted as  
$\tilde{X} \in \mathbb{R}^{S \times HW}$.  
The observation model can then be formulated as:
\begin{equation}
Y \approx \tilde{X}BS + N_y, \quad Z \approx R\tilde{X} + N_z,
\end{equation}
where $B$, $S$, and $R$ denote the blurring, downsampling, and spectral response operators, respectively. Although $\tilde{X}$ is unobservable, the reconstructed image $X$ is expected to approximate it. Following this degradation model, we first spatially align $Y$ and $Z$ and extract shallow features via a cross-modal fusion module:
\begin{equation}
\begin{aligned}
Y_{up} &= U_p(Y), \\
(F'_y &, F'_z) = CMFB(Cat( F_{conv}(Z), F_{conv}(Y_{up}))), \\
F_{spe} &= \sigma(\alpha) \odot F_{conv}(Y_{up}) + \sigma(\beta) \odot F'_y, \\
F_{spa} &= \sigma(\gamma) \odot F_{conv}(Z) + \sigma(\zeta) \odot F'_z,
\end{aligned} 
\end{equation} 
where $U_p(\cdot)$ denotes the upsampling operation, $\mathrm{Cat}(\cdot)$ represents channel concatenation, and $\mathrm{CMFB}(\cdot)$ refers to the cross-modal fusion. Furthermore, $\sigma(\cdot)$ denotes the sigmoid activation function, while $\odot$ signifies element-wise multiplication. \par

To address the limitations of existing fusion methods in modeling spatial topology and pixel-level feature manifold relationships, we propose a Topology-Aware Transformer (TPFormer) that integrates topology-aware attention with neighborhood propagation to jointly model spatial topology and pixel-level feature manifold relationships, thereby capturing intrinsic spatial–spectral structures and improving topology-aware representation learning. The extracted shallow features $F_{spe}$ and $F_{spa}$ are subsequently injected into the TPFormer module to capture fine-grained feature representations, with the overall architecture illustrated in Fig. \ref{fig:tpformer}. For the LR-HSI branch, the graph construction mapping is defined as $\Phi: \mathbb{R}^{C \times H \times W} \to G$, where $\Phi(X) = (V, E)$. Here, $V$ denotes the node feature matrix, and $E$ represents the set of graph edges. Two complementary graphs are constructed: (1) a spatial adjacency graph based on image topology and (2) a feature similarity graph constructed from pixel spectral-aware latent features using K-nearest neighbors.
\begin{equation}
\begin{aligned}
\mathcal{E}&_{\mathrm{spec}} = KNN(F, K=k_{\mathrm{spec}}), \\
\mathcal{E}&_{\mathrm{spat}} = SpatialEdges(H, W),
\end{aligned}
\end{equation}
where $F= reshape\left(F_{spe}^{patch},(C, H \cdot W)\right)^T \in \mathbb{R}^{N \times C}$,  $k_{\mathrm{spec}}$ denotes a constant.  $\mathcal{E}_{\mathrm{spec}}$ denotes the set of edge indices corresponding to the feature manifold graph, $KNN(\cdot)$ identifies the k nearest neighbors of each pixel based on feature similarity, thereby establishing the graph's edge topology. 
Likewise,  $\mathcal{E}_{\mathrm{spat}}$ represents the spatial 4-neighborhood graph. \par
Although Transformer self-attention captures dense global contextual interactions, it primarily learns pairwise correlations based on feature similarity without explicitly modeling the geometric structure of the underlying manifold. Consequently, attention weights do not guarantee neighborhood consistency or topology-preserving information propagation. Unlike self-attention that captures semantic affinity, graph propagation explicitly encodes local manifold structural by constraining information exchange within topology-consistent neighborhoods. Therefore, GNN propagation preserves structural continuity of the underlying manifold during representation learning. To further enlarge the capable graph receptive field, Chebyshev polynomial expansion is employed to construct a high-order graph diffusion operator, defined as:
\begin{equation}
\begin{aligned}
D(E) &= \{(i, j) \mid A_{\mathrm{diff}}[i, j] > \tau\}, \\
A_{\mathrm{diff}} &= \sum_{k=0}^{K} \alpha_k T_k(\tilde{\mathcal{L}})\\
\mathcal{G} = (V&, E_{\mathrm{diff}}), \quad E_{\mathrm{diff}} = D(\mathcal{E}_{\mathrm{spec}} \cup \mathcal{E}_{\mathrm{spat}}),
\end{aligned}
\end{equation}
where $\mathcal{L} = I-\tilde{D}^{-1/2} (A + I) \tilde{D}^{-1/2}$ denotes the normalized Laplacian operator, $T_k(\cdot)$ denotes the $k$-th order Chebyshev polynomial. The symbol $\mathcal{G}$ denotes the final output graph structure. 
Based on the constructed diffusion graph, TPFormer employs three graph operators with complementary neighborhood aggregation mechanisms to perform topology-aware message propagation. Specifically, GCN captures smooth manifold continuity through isotropic aggregation, GAT adaptively emphasizes informative neighbors via attention weighting, and SuperGAT further enhances topology consistency through self-supervised edge regularization. Their complementary aggregation mechanisms produce diverse manifold representations, which are adaptively fused to improve topology-aware spatial–spectral representation learning.
The graph neural network mapping is defined as $\Psi_{\theta}: \mathcal{G} \rightarrow \mathbb{R}^{N \times C}$, with $\Psi_{\theta}(\mathcal{G}) = H$. The feature representations extracted from the three branches are denoted as
\begin{equation}
\begin{aligned}
H_{\mathrm{gat}} &={GAT}_2 \circ {GAT}_1( \mathcal{G}), \\
H_{\mathrm{gcn}} &= {GCN}_2 \circ {GCN}_1( \mathcal{G}), \\
H_{\mathrm{super}} &= {SuperGAT}_2 \circ {SuperGAT}_1( \mathcal{G}), \\
\end{aligned}
\end{equation}
where $\mathrm{GAT}_k(\mathcal{G})$, $\mathrm{GCN}_k(\mathcal{G})$, and $\mathrm{SuperGAT}_k(\mathcal{G})$ denote graph attention, graph convolution, and edge-aware graph attention operations, respectively, which provide complementary topology-aware feature propagation. $\circ$ denotes function composition. The fusion function is defined as:
\begin{equation}
H_{\mathrm{fused}} = \mathcal{F}(H_{\mathrm{gat}}, H_{\mathrm{gcn}}, H_{\mathrm{super}}) = \sum_{i=1}^{3} w_i H_i,
\end{equation}
where $ {w} = [w_1, w_2, w_3]^T \in \Delta^2 $ is a learnable weight vector.\par
We design a bidirectional attention mechanism and formalize the attention function as a spatial mapping $\mathcal{T}: \mathbb{R}^{N \times C} \times \mathbb{R}^{N \times C} \to \mathbb{R}^{N \times C}$
\begin{equation}
\begin{aligned}
\mathcal{T}(Q, K, V) = \mathrm{softmax}\!\left( \frac{Q K^T}{\sqrt{d}} \right) V, \\
\mathcal{T} = \alpha \cdot \mathcal{T}(H_{\mathrm{fused}}, \mathcal{V}, \mathcal{V}) + (1 - \alpha) \cdot \mathcal{T}(\mathcal{V}, H_{\mathrm{fused}}, H_{\mathrm{fused}}).
\end{aligned}
\end{equation}
where $\alpha = \sigma(\alpha')$ denotes a learnable attention parameter. The normalization function $\mathcal{N}: \mathbb{R}^{N \times C} \to \mathbb{R}^{N \times C}$ is implemented via layer normalization, i.e., $\mathcal{N}({X}) = \text{LayerNorm}({X})$. The feed-forward network $\mathcal{F}_{\mathrm{ffn}}: \mathbb{R}^{N \times C} \to \mathbb{R}^{N \times C}$ is formulated as $
\mathcal{F}_{\mathrm{ffn}}({X}) = \text{GELU}({X} {W}_1) {W}_2.$ Meanwhile, residual connections are applied as follows:
${R}_1 = \mathcal{N}({H}_{\mathrm{fused}} + \mathcal{T}), \quad
{R}_2 = \mathcal{N}({R}_1 + \mathcal{F}_{\mathrm{ffn}}({R}_1))$. The output reconstruction mapping $\mathcal{I}: \mathbb{R}^{N \times C} \to \mathbb{R}^{C \times H \times W}$ is defined as $\mathcal{I}({R}_2) = \mathrm{reshape}({R}_2)$, which transforms the flattened feature matrix back into a three-dimensional tensor. The output from the HR-MSI branch is obtained analogously.

\subsection{Frequency-Decoupled Spatial-Spectral Collaborative Fusion}
Existing fusion strategies usually process spatial–spectral features in an entangled manner, which inevitably introduces low-frequency bias and weakens the reconstruction of high-frequency geometric details. To address this issue, FDSCF explicitly decouples spatial–spectral representations into low- and high-frequency components and performs frequency-specific feature refinement, enabling robust preservation of structural details and spectral fidelity.
\paragraph{Spectral-Driven Attention Fusion}

Let ${X}{\mathrm{msi}} \in \mathcal{M}{\mathrm{MSI}}$ and ${X}{\mathrm{hsi}} \in \mathcal{M}{\mathrm{HSI}}$ denote the TPFormer features of the HR-MSI and LR-HSI branches, respectively. A Spectral-Driven Attention Fusion (SDAF) module is introduced to establish cross-modal spectral correspondence. SDAF adopts a Transformer-based architecture with cross-attention to align spectral responses between modalities while preserving spatial coherence through residual connections and positional embeddings. This operation produces spectrally aligned MSI features $F_{\mathrm{out}}^{\mathrm{msi}}$ that serve as a reliable high-resolution spatial prior. The detailed structure outlined below:
\begin{equation}
\begin{aligned}
A &= \mathrm{SpectralMSA}(\hat{X}_{\mathrm{msi}}, \hat{X}_{\mathrm{hsi}}) + \hat{X}_{\mathrm{msi}}, \\
Z &= \mathcal{P}^{-1}(A), \\
F_{\mathrm{out}}^{\mathrm{msi}} &= (\mathrm{FFN} + \mathcal{I}) \circ \mathcal{P}^{-1} \left( \mathrm{LN} \left( A + \mathcal{P} \left( \mathrm{CNN}_{\mathrm{pos}} (Z) \right) \right) \right).
\end{aligned}
\end{equation}
where the normalized and permuted inputs are defined as $\hat{{X}}_{{msi}} = {LN}_{{msi}}(\mathcal{P}({X}_{\mathrm{msi}}))$ and $\hat{{X}}_{\mathrm{hsi}} = {LN}_{{hsi}}(\mathcal{P}({X}_{\mathrm{hsi}}))$, with $\mathcal{P} = \mathcal{P}_{(0,3,2,1)}$ denoting the tensor permutation operator that reorders the dimensions from $(N, C, H, W)$ to $(N, W, H, C)$, effectively moving the channel dimension to the last position and transposing the spatial dimensions. The inverse permutation $\mathcal{P}^{-1}$ restores the original layout. In addition, LN refers to the LayerNorm operation, $CNN_{pos}$ denotes the positional embedding module, consisting of two convolutional layers followed by a GELU activation, FFN represents the feed-forward network, and SpectralMSA(·) denotes the cross-modal attention fusion module.\par
\paragraph{Spectral-Driven Spatial Enhancement}
To enhance spatial representations under spectral guidance, we propose a Spectral-Driven Spatial Enhancement (SDSE) module. SDSE combines low-rank spatial projections with linear attention to capture long-range dependencies efficiently, while convolutional positional embeddings exploit local spatial structures. A low-rank guided spatial injection mechanism further refines spatial and channel responses, producing structurally enhanced and spectrally consistent features. Specifically, two low-rank transformation functions are defined as:
$L_w: \mathbb{R}^{B \times C \times H \times W} \rightarrow \mathbb{R}^{B \times C \times H \times w'}$ and
$L_h: \mathbb{R}^{B \times C \times H \times W} \rightarrow \mathbb{R}^{B \times C \times h' \times W}$.
They are implemented using depthwise separable convolutions along the width and height dimensions with kernel sizes of $(1, L)$ and $(L, 1)$, respectively, which perform low-rank compression while projecting the input features into the attention space:
\begin{equation}
\begin{aligned}
    s_{\mathrm{msi}} &= L_w(X_{\mathrm{msi}}), 
    a_{\mathrm{msi}} = L_h(X_{\mathrm{msi}}), \\
    \tilde{s}_{\mathrm{msi}}^{(d)} &= \Phi_{\mathrm{up}}(s_{\mathrm{msi}}),  \tilde{a}_{\mathrm{msi}}^{(d)} = \Phi_{\mathrm{up}}(a_{\mathrm{msi}}), \\
    \mathcal{A}_{\mathrm{lin}}(\tilde{s}_{\mathrm{msi}}^{(d)}, \tilde{a}_{\mathrm{msi}}^{(d)}) &= \Phi_{\mathrm{out}} \left( \text{LinearAttention} \left( \tilde{s}_{\mathrm{msi}}^{(d)}, \tilde{a}_{\mathrm{msi}}^{(d)} \right) \right),
\end{aligned}
\end{equation}
where $\Phi_{\mathrm{up}}:\mathbb{R}^{C}\rightarrow\mathbb{R}^{d}$ denotes a linear up-projection, with $d=\text{heads}\times\text{head\_dim}$, and $\mathrm{LinearAttention}(\cdot)$ denotes the linear attention operator. To exploit spatial coherence and geometric consistency, convolutional positional embeddings are introduced as follows: 
\begin{equation}
\mathcal{P}_{\mathrm{emb}} = \mathrm{CNN}_{\mathrm{pos}} \circ \Phi_{\mathrm{up}} \circ \mathcal{A}_{\mathrm{lin}}(\tilde{s}_{\mathrm{msi}}^{(d)},  \tilde{a}_{\mathrm{msi}}^{(d)})
\end{equation}
where $\mathrm{CNN}_{\mathrm{pos}}(\cdot)$ denotes a convolutional positional embedding for preserving local spatial consistency. A low-rank feature fusion module is further introduced to jointly enhance spatial and channel representations via low-rank guided spatial injection. The fusion process is formulated as follows:
\begin{equation}
\begin{aligned}
    \mathcal{A}_{\mathrm{lowrank}}(\mathcal{X}) &= \mathcal{W}_o * \mathrm{Att}\!\left( \mathcal{W}_q * \mathcal{X},\, \mathcal{W}_k * \mathcal{X},\, \mathcal{W}_v * \mathcal{X} \right), \\
    \mathrm{Att}(Q,K,V) &= \mathrm{Reshape}\!\left( V_f \cdot \sigma(Q_f^\top K_f / \sqrt{r}),\, (B,C,H,W) \right),
\end{aligned}
\end{equation}
where $Q_f, K_f, V_f$ denote the flattened query, key, and value projections, respectively, and $\sigma(\cdot)$ represents the softmax operation. Finally, the attention output, positional encoding, and low-rank attention output are fused to produce the final output of the SDSE module, as follows:
\begin{equation}
\begin{aligned}
    \mathcal{Y} &= \mathcal{A}_{\mathrm{lin}} + \mathcal{P}_{\mathrm{emb}} + \mathcal{A}_{\mathrm{lowrank}}(\mathcal{X}_{\mathrm{hsi}}), \\
    F_{\mathrm{out}}^{\mathrm{hsi}} &=  \mathcal{Y} + \text{FFN}( \mathcal{Y}).
\end{aligned}
\end{equation}\par
\begin{figure*}[htbp]    
\centering     
\includegraphics[width=1.0\textwidth]{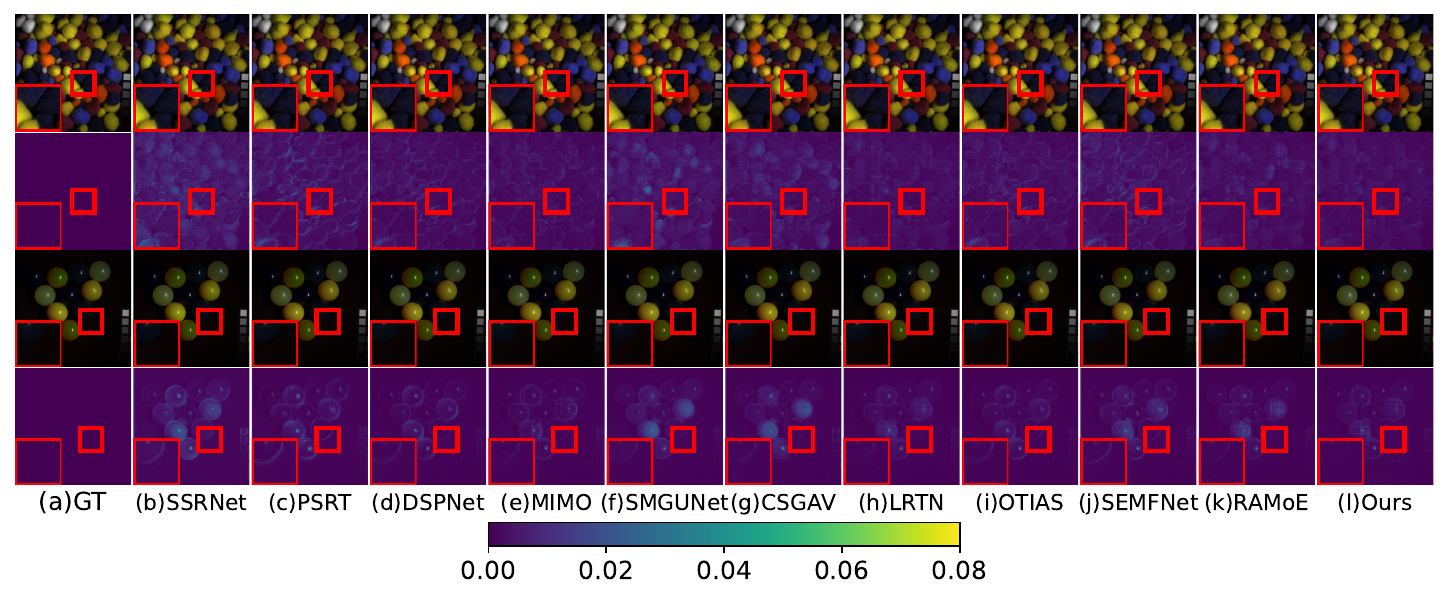}
\caption{A comparative visualization of different methods on the CAVE dataset, with residual plots showing the differences between the reconstructed images and the ground truth (GT).}
\end{figure*}
\paragraph{Frequency-Decoupled Enhancement via DCT}
Although attention-based fusion effectively aggregates contextual information, the resulting features remain dominated by low-frequency components, leading to blurred edges and structural degradation. To alleviate this issue, FDSCF performs frequency-decoupled enhancement via the Discrete Cosine Transform (DCT). Specifically, the outputs $F_{\mathrm{out}}^{\mathrm{msi}}$ and $F_{\mathrm{out}}^{\mathrm{hsi}}$ are first transformed into the frequency domain via a separable DCT operator applied independently across channels:
\begin{equation}
F_{\mathrm{out}}^{\mathrm{msi}},\, F_{\mathrm{out}}^{\mathrm{hsi}} = \mathcal{D}\!\left(F_{\mathrm{out}}^{\mathrm{msi}},\, F_{\mathrm{out}}^{\mathrm{hsi}}\right),
\end{equation}
where $\mathcal{D}(\cdot)$ denotes the channel-wise DCT transform.
A binary low-frequency mask ${M}_L \in \{0,1\}^{H \times W}$ is defined by $[{M}_L]_{i,j} = \mathbb{I}(i < L \land j < L)$, with $L$ a hyperparameter and $\mathbb{I}(\cdot)$ the indicator function. The high-frequency coefficients are then explicitly isolated through spectral masking:
     \begin{align}
    {F}_{\mathrm{out}}^{\mathrm{msi}} &= {F}_{\mathrm{out}}^{\mathrm{msi}} \odot ({M} - {M}_L), \\
    {F}_{\mathrm{out}}^{\mathrm{hsi}} &= {F}_{\mathrm{out}}^{\mathrm{hsi}} \odot ({M} - {M}_L),
\end{align}
where $\odot$ denotes element-wise multiplication and ${M}$ is an all-ones matrix. The extracted high-frequency coefficients are reconstructed into the spatial domain using the inverse DCT:
 \begin{equation}
{F}_{\mathrm{out}}^{\mathrm{msi}},{F}_{\mathrm{out}}^{\mathrm{hsi}} = \mathcal{D}^{-1}({F}_{\mathrm{out}}^{\mathrm{msi}},{F}_{\mathrm{out}}^{\mathrm{hsi}} )
 \end{equation} \par
\paragraph{Spatial–Spectral Collaborative Fusion}
Following frequency-domain enhancement, a Channel-Spatial Attention (CSA) module is employed to adaptively integrate the spatial and spectral branches. Residual connections recover the original feature information while incorporating enhanced high-frequency details. Consequently, the fused representations simultaneously retain spectral fidelity and structural details. The details are as outlined below:
\begin{equation}
\begin{aligned}
    x_{\mathrm{out}}^{\mathrm{msi}},\, x_{\mathrm{out}}^{\mathrm{hsi}} &= \mathrm{CSA}\left(F_{\mathrm{out}}^{\mathrm{msi}},\, F_{\mathrm{out}}^{\mathrm{hsi}}\right), \\
    X_{\mathrm{msi}} &= x_{\mathrm{out}}^{\mathrm{msi}} + F_{\mathrm{out}}^{\mathrm{msi}}, \\
    X_{\mathrm{hsi}} &= F_{\mathrm{out}}^{\mathrm{hsi}} + x_{\mathrm{out}}^{\mathrm{hsi}},
\end{aligned}
\end{equation}
where $\mathrm{CSA}(\cdot)$ represents the channel-spatial cross-attention mechanism. The fused features are subsequently reconstructed into the high-resolution hyperspectral image. \par
Low-resolution hyperspectral images are fused with high-resolution multispectral images to generate HR-HSI. To achieve high-fidelity reconstruction, the reconstruction loss is formulated using the $\mathcal{L}_1$ loss function and is given by:
\begin{equation}
    \mathcal{L}_{\mathrm{rec}}(\hat{X}, X) = \|\hat{X} - X\|_1,
\end{equation}
where $X$ and $\hat{X}$ denote the reference and reconstructed high-resolution hyperspectral images, respectively.\par
\begin{figure*}[htbp]    
\centering     
\includegraphics[width=1.0\textwidth]{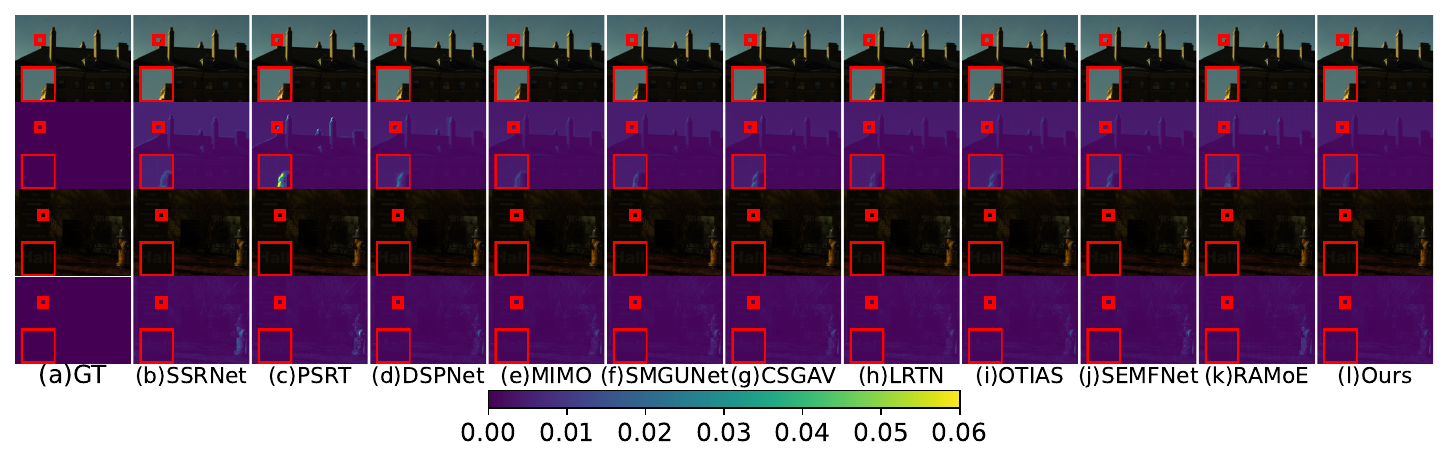}
\caption{A comparative visualization of different methods on the Harvard dataset, with residual plots showing the differences between the reconstructed images and the GT.}
\end{figure*}
\begin{figure*}[htbp]    
\centering     
\includegraphics[width=1.0\textwidth]{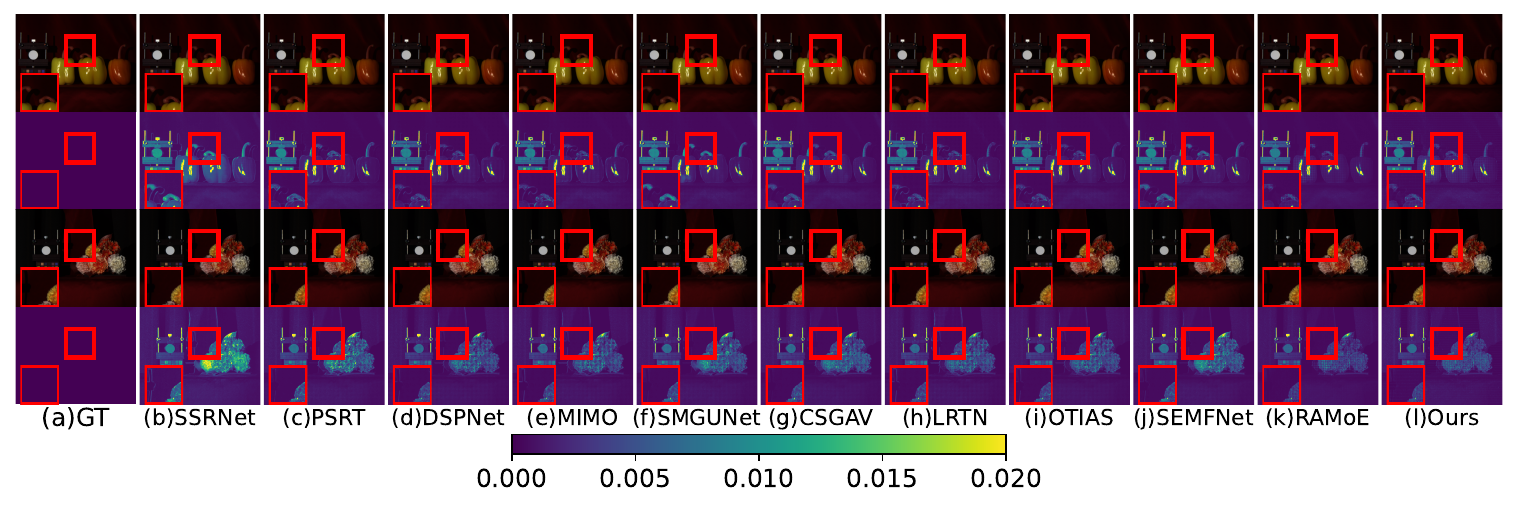}
\caption{ A comparative visualization of different methods on the KAIST dataset, with residual plots showing the differences between the reconstructed images and the GT.}
\end{figure*}
\section{Experiment}
\subsection{Experiment settings}
\textbf{Datasets:} The proposed model is evaluated on three public benchmark datasets, namely CAVE, Harvard, and KAIST. The CAVE dataset contains high-resolution hyperspectral images of natural and synthetic materials with 31 spectral bands spanning 400–700 nm. The Harvard dataset consists of hyperspectral scenes acquired from diverse indoor and outdoor environments, each with a spatial resolution of $1040\times1392$ pixels. The KAIST dataset provides hyperspectral data for analyzing surface material characteristics under diverse conditions. For CAVE and KAIST, the first 20 samples are used for training and the remaining samples for testing. In the case of the Harvard dataset, the initial 30 hyperspectral samples are utilized for training, and the others are allocated for testing.\par
In real-world imaging scenarios, acquiring HR-HSIs is challenging due to sensor limitations and low signal-to-noise ratios. To generate training data that reflect realistic degradation processes, a simulation-based degradation model is adopted to synthesize LR-HSIs and HR-MSIs. Specifically, LR-HSIs are obtained by convolving the original HR-HSIs with an $8\times 8$ Gaussian kernel of standard deviation 3, followed by an eight-fold spatial downsampling operation. HR-MSIs are generated through spectral resampling using the spectral response functions of the Nikon D700 camera. During preprocessing, local patches are extracted using a sliding-window strategy to enhance spatial diversity in the training set. For the CAVE dataset, $64\times64$ patches are extracted with a stride of 32, whereas a stride of 64 is adopted for the Harvard and KAIST datasets to reduce redundancy and improve efficiency.
\par

\textbf{Parameter Setting:} All experiments were conducted on an NVIDIA RTX 4090 GPU using PyTorch 2.4.0. The model was optimized using Adam with $\beta_1=0.9$ and $\beta_2=0.999$. The initial learning rate was set to $4 \times 10^{-4}$, with decay controlled by a cosine annealing schedule during iterative optimization. A batch size of 16 was used consistently across all datasets. During graph construction, four-order Chebyshev polynomials were employed to capture long-range node dependencies, and all graph convolutional networks used default settings. The number of FDSCF blocks is set to 2.\par
\subsection{Compared methods}
To assess the effectiveness of the proposed model, five quantitative evaluation metrics are utilized: the Peak Signal-to-Noise Ratio (PSNR) to measure pixel-level reconstruction fidelity, the Spectral Angle Mapper (SAM) to evaluate spectral similarity, the Erreur Relative Globale Adimensionnelle de Synthèse (ERGAS) to quantify overall relative error, the Root Mean Square Error (RMSE) to indicate average reconstruction deviation, and the Universal Image Quality Index (UIQI). Comparative analyses are performed on three publicly accessible datasets, where the proposed approach is benchmarked against ten state-of-the-art methods, namely SSRNet \cite{9186332}, PSRT \cite{10044141}, DSPNet \cite{10264151}, MIMO \cite{10419133}, LRTN \cite{liu2025low}, SMGUNet \cite{YAN2025111277}, CSGAV \cite{11014272}, OTIAS \cite{deng2025otias}, SEMFNet \cite{11347025}, and RAMoE \cite{Xiao2026RAMoE}.
The experimental outcomes demonstrate that the designed model consistently surpasses existing techniques across most quantitative indicators, confirming its superior reconstruction accuracy.
\begin{table*}
    \centering
    \caption{Experimental comparison of different methods on the CAVE, Harvard, and KAIST datasets. Optimal and suboptimal values are indicated in bold and underline, respectively.}
    \label{tab:quantitative_comparison}
    \renewcommand{\arraystretch}{0.8} 
    \begin{tabular*}{0.95\textwidth}{@{\extracolsep{\fill}} l c c c c c c} 
        \toprule
        \multirow{2}{*}{\textbf{Dataset}} & \multirow{2}{*}{\textbf{Model}} & \multicolumn{5}{c}{\textbf{Metrics}} \\
        \cmidrule(lr){3-7}
         & & \textbf{PSNR} & \textbf{SAM} & \textbf{ERGAS} & \textbf{RMSE} & \textbf{UIQI} \\
        \midrule
        
        \multirow{10}{*}{\textbf{CAVE}} & SSRNet \cite{9186332} & 43.6021 & 4.4768 & 0.9048 & 2.0789 & 0.9849 \\
         & PSRT  \cite{10044141} & 45.9228 & 3.1022 & 0.7010 & 1.6614 & 0.9898 \\
         & DSPNet \cite{10264151} & 47.6970 & 2.4971 & 0.5800 & 1.3282 & 0.9926 \\
         & MIMO \cite{10419133}  & 48.2850 & 2.5157 &  0.5477 & 1.2588 & 0.9924 \\
         & SMGUNet \cite{YAN2025111277} & 46.2577 & 2.8722 & 0.7831 & 1.7826 & 0.9912 \\
         & CSGAV \cite{11014272} & 46.8428 & 2.5950 & 0.6600 & 1.4939 & 0.9921 \\
         & LRTN \cite{liu2025low} & 48.1070 &  2.3358 & 0.5517 &  \underline{1.2524} & \underline{0.9929} \\
         & OTIAS \cite{deng2025otias} & 48.0643 & 2.3984 & 0.5555 & 1.2919 & \underline{0.9929} \\
         & SEMFNet \cite{11347025} & 46.3941 & 2.8798 & 0.6647 & 1.5290 & 0.9914 \\
         & RAMoE \cite{Xiao2026RAMoE} & \underline{48.2987} & \underline{2.2544} & \underline{0.5415} & 1.2546 & \textbf{0.9939} \\
         & \textbf{Ours} & \textbf{48.9392} & \textbf{2.1879} & \textbf{0.5069} & \textbf{1.1393} & \textbf{0.9939} \\
        \midrule
        
        \multirow{10}{*}{\textbf{Harvard}} & SSRNet \cite{9186332} & 46.2318 & 3.2650 & 1.1660 & 1.9176 & 0.9679 \\
        & PSRT  \cite{10044141} & 45.9949 & 3.1416 & 1.1960 & 1.9284 & \underline{0.9705} \\
        & DSPNet \cite{10264151} & 47.1961 & 2.8715 & 1.0206 & 1.7389 & 0.9692 \\
        & MIMO \cite{10419133} & 47.2452 & 2.9163 & 1.0241 & 1.7160 & 0.9687 \\
        & SMGUNet \cite{YAN2025111277} & 47.0914 & 2.8851 & 1.0518 & 1.7484 & 0.9699 \\
        & CSGAV \cite{11014272} & 47.1157 & 2.8730 & 1.0376 & 1.7524 & 0.9697 \\
        & LRTN \cite{liu2025low} & 47.2064 &  \underline{2.8504} & 1.0108 & 1.7286 & \textbf{0.9707} \\
        & OTIAS \cite{deng2025otias} &  \underline{47.3361} & 2.8553 &  \underline{0.9972} &  \underline{1.7146} & 0.9698 \\
        &  SEMFNet \cite{11347025} & 47.0949 & 2.9150 & 1.0537 & 1.7731 & 0.9688 \\
        &  RAMoE \cite{Xiao2026RAMoE} & 46.9968 & 3.0722 & 1.0589 & 1.7692 & 0.9678 \\
        & \textbf{Ours} & \textbf{47.4452} & \textbf{2.8312} & \textbf{0.9802} & \textbf{1.6934} & 0.9699 \\
        \midrule
        
        \multirow{9}{*}{\textbf{KAIST}} & SSRNet \cite{9186332} & 43.7750 & 3.6042 & 1.3590 & 1.7766 & 0.9849 \\
        & PSRT \cite{10044141} & 44.9084 & 2.7289 & 1.2006 & 1.5291 & 0.9875 \\
        & DSPNet \cite{10264151} & 45.7651 &  \underline{2.4334} & 1.0888 & 1.3985 & 0.9887 \\
        & MIMO \cite{10419133}  & 45.8058 & 2.4903 & 1.0846 & 1.3871 & 0.9884 \\
        & SMGUNet \cite{YAN2025111277} & 45.4432 & 2.5738 & 1.1272 & 1.4443 & 0.9879 \\
        & CSGAV \cite{11014272} & 45.5340 & 2.5409 & 1.1159 & 1.4266 & 0.9879 \\
        & LRTN \cite{liu2025low} & 45.8611 & 2.4585 & 1.0756 & 1.3777 & 0.9886 \\
        & OTIAS \cite{deng2025otias} &  45.9094 & 2.4492 & 1.0708 &  1.3717 & \underline{0.9888} \\
        & SEMFNet \cite{11347025} & 45.0100 & 2.6854 & 1.2715 & 1.5130 & 0.9872 \\
        & RAMoE \cite{Xiao2026RAMoE} &  \underline{45.9589} & 2.4570 &  \underline{1.0659} & \underline{1.3649} & 0.9885 \\
        & \textbf{Ours} & \textbf{46.0497} & \textbf{2.4158} & \textbf{1.0647} & \textbf{1.3525} & \textbf{0.9889} \\
        \bottomrule
    \end{tabular*}
\end{table*}
\begin{table}
\centering
\caption{Complexity comparison for different models.}
\label{tab:FLOPs and Parameters}
\renewcommand{\arraystretch}{0.7} 
\begin{tabular*}{0.45\textwidth}{@{\extracolsep{\fill}} l c c } 
\toprule
\textbf{Model} & \textbf{FLOPs (G)} & \textbf{Params (M)} \\
\midrule
SSRNet \cite{9186332}         & 0.213             & 0.026             \\
PSRT \cite{10044141}          & 2.101              & 0.247             \\
DSPNet \cite{10264151}        & 13.194             & 6.055             \\
MIMO \cite{10419133}          & 3.068              & 1.622             \\
SMGUNet \cite{YAN2025111277}          & 11.714             & 0.759             \\
CSGAV \cite{11014272}          & 8.915              & 1.271             \\
LRTN \cite{liu2025low}          & 4.143              & 3.535             \\
OTIAS \cite{deng2025otias}          & 15.531             & 2.924             \\
SEMFNet \cite{11347025}       & 8.615             & 1.002             \\
RAMoE \cite{Xiao2026RAMoE}       & 230.642             & 28.170             \\
Ours           & 6.521            & 3.933             \\
\bottomrule
\end{tabular*}
\end{table}
\subsection{Results of Hyperspectral Image Fusion}
The effectiveness of the proposed DDMM framework is validated on the CAVE, Harvard, and KAIST datasets through comprehensive quantitative and qualitative evaluations. As reported in Table \ref{tab:quantitative_comparison}, DDMM demonstrates superior performance across the majority of metrics on all datasets, demonstrating superior spatial reconstruction accuracy and spectral fidelity. These consistent improvements over competing methods demonstrate the robustness and generalization capability of the proposed framework under diverse imaging conditions. Table \ref{tab:FLOPs and Parameters} summarizes the computational complexity of different methods. Despite its moderate computational cost, DDMM achieves superior reconstruction performance while maintaining competitive efficiency, demonstrating a favorable trade-off between reconstruction quality and model complexity. Representative reconstruction results and corresponding residual maps are presented in Figs. 3-5. Compared with competing approaches, DDMM reconstructs more accurate structural details while producing fewer artifacts and lower residual errors relative to the ground truth. These results further confirm the effectiveness of the proposed framework in preserving spatial–spectral information for hyperspectral image reconstruction.\par

\begin{figure}[htbp]    
\centering     
\includegraphics[width=1.0\columnwidth]{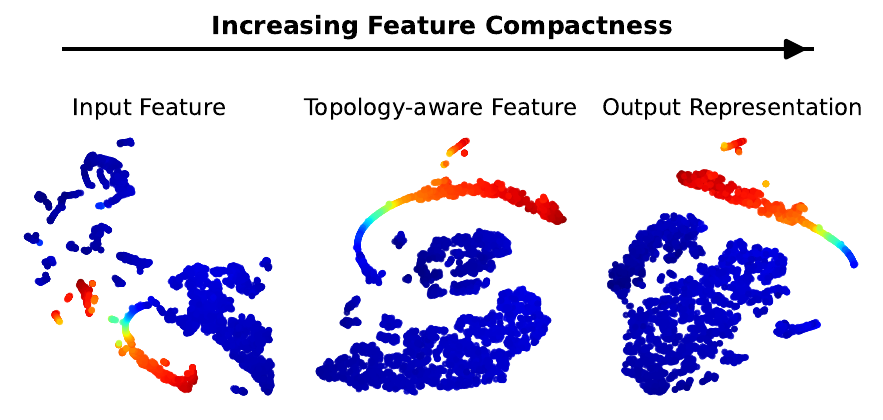}
\caption{Visualization of feature structural refinement induced by topology-aware propagation in TPFormer. Feature embeddings are visualized using t-SNE, where colors are assigned according to the normalized first principal component of the original spectral signatures to reflect continuous spectral variations across pixels.}
\end{figure}

\begin{table}[htbp]
\centering
\caption{Ablation study results on the CAVE dataset.}
\label{tab:ablation_cave}
\scriptsize 
\setlength{\tabcolsep}{7pt} 
\renewcommand{\arraystretch}{1.1}
\begin{tabular}{l c c c c c}
\toprule
\textbf{Model} 
 & \textbf{PSNR} & \textbf{SAM} & \textbf{RMSE} & \textbf{ERGAS} &\textbf{UIQI} \\
\midrule
w/o All & 47.2060 & 2.5539 & 1.4146 & 1.3648 & 0.9887 \\
\hline
w/o Chebyshev & 48.6308 & 2.2394 & 1.2094 & 0.5355 &  0.9931 \\
w/o GNN & 48.4371 & 2.3036 & 1.2847 & 0.5459 &  0.9928 \\
w/o TPFormer & 48.4996 & 2.2302 & 1.2154 & 0.5304 & 0.9937 \\
\hline
w/o SDAF & 48.7741 & 2.2468  & 1.1726 & 0.5205  &   0.9938  \\
w/o SDSE & 48.5628 & 2.3607  & 1.2622 &  0.5441  &  0.9930 \\
w/o Lowrank & 48.7019 & 2.2322  & 1.1774 & 0.5290  &  0.9937  \\
w/o DCT & 48.1973 & 2.3965  & 1.2684 & 0.5568  &   0.9929  \\
w/o FDSCF & 47.8723  &2.6969  &1.3027  &0.5694 & 0.9922 \\
\hline

Ours & \textbf{48.9392} & \textbf{2.1879} & \textbf{1.1393} &\textbf{0.5069} & \textbf{0.9939}  \\
\bottomrule
\end{tabular}
\end{table}
\begin{table}[t]
\centering
\caption{Graph operator analysis on the CAVE dataset.}
\label{tab:Graph_cave}
\scriptsize 
\setlength{\tabcolsep}{6pt} 
\renewcommand{\arraystretch}{1.1}
\begin{tabular}{l c c c c c}
\toprule
\textbf{Model} 
 & \textbf{PSNR} & \textbf{SAM} & \textbf{RMSE} & \textbf{ERGAS} &\textbf{UIQI} \\
\midrule
Transformer & 48.4371 & 2.3036 & 1.2847 & 0.5459 &  0.9928 \\
Transformer+GCN & 48.6928 & 2.2373 & 1.1685 & 0.5162 & 0.9937   \\
Transformer+GAT & 48.7549 & 2.2269 & 1.1670 & 0.5120 & 0.9937 \\
Transformer+SuperGAT & 48.7924  &  2.2194  & 1.1523  & 0.5096 &  0.9938 \\
Transformer+GCN+GAT & 48.8420  &  2.2049  & 1.1524  & 0.5087 &  0.9938 \\
Ours & \textbf{48.9392} & \textbf{2.1879} & \textbf{1.1393} &\textbf{0.5069} & \textbf{0.9939}  \\
\bottomrule
\end{tabular}
\end{table}

\subsection{Ablation study}
To validate the effectiveness of DDMM, ablation studies are conducted on the CAVE dataset by systematically removing key components. The quantitative results are reported in Table~\ref{tab:ablation_cave}. Specifically, ``w/o TPFormer'', ``w/o FDSCF'', and ``w/o All'' denote removing TPFormer, FDSCF, and both modules, respectively. ``w/o Chebyshev'' disables high-order topology propagation by removing the Chebyshev polynomial approximation, while ``w/o GNN'' removes all graph neural network branches (GCN, GAT, and SuperGAT). ``w/o SDAF'', ``w/o SDSE'', and ``w/o Lowrank'' respectively remove the spectral-driven attention, spatial enhancement, and low-rank structural prior in FDSCF. ``w/o DCT'' removes the DCT-based frequency modeling branch, eliminating global frequency-domain representations and their complementary interaction with spatial features. All variants are trained and evaluated under identical experimental settings. As shown in Table~\ref{tab:ablation_cave}, removing any individual component consistently degrades reconstruction performance. The inferior results of ``w/o GNN'' and ``w/o Chebyshev'' highlight the importance of topology-aware graph propagation, while the degradation of ``w/o DCT'' demonstrates the benefit of frequency-domain modeling. In addition, the performance declines of ``w/o SDAF'', ``w/o SDSE'', and ``w/o Lowrank'' verify the effectiveness of spectral-driven attention, spatial enhancement, and low-rank  structural priors in frequency-decoupled spatial--spectral fusion.

\begin{figure*}[htbp]    
\centering     
\includegraphics[width=0.9\textwidth]{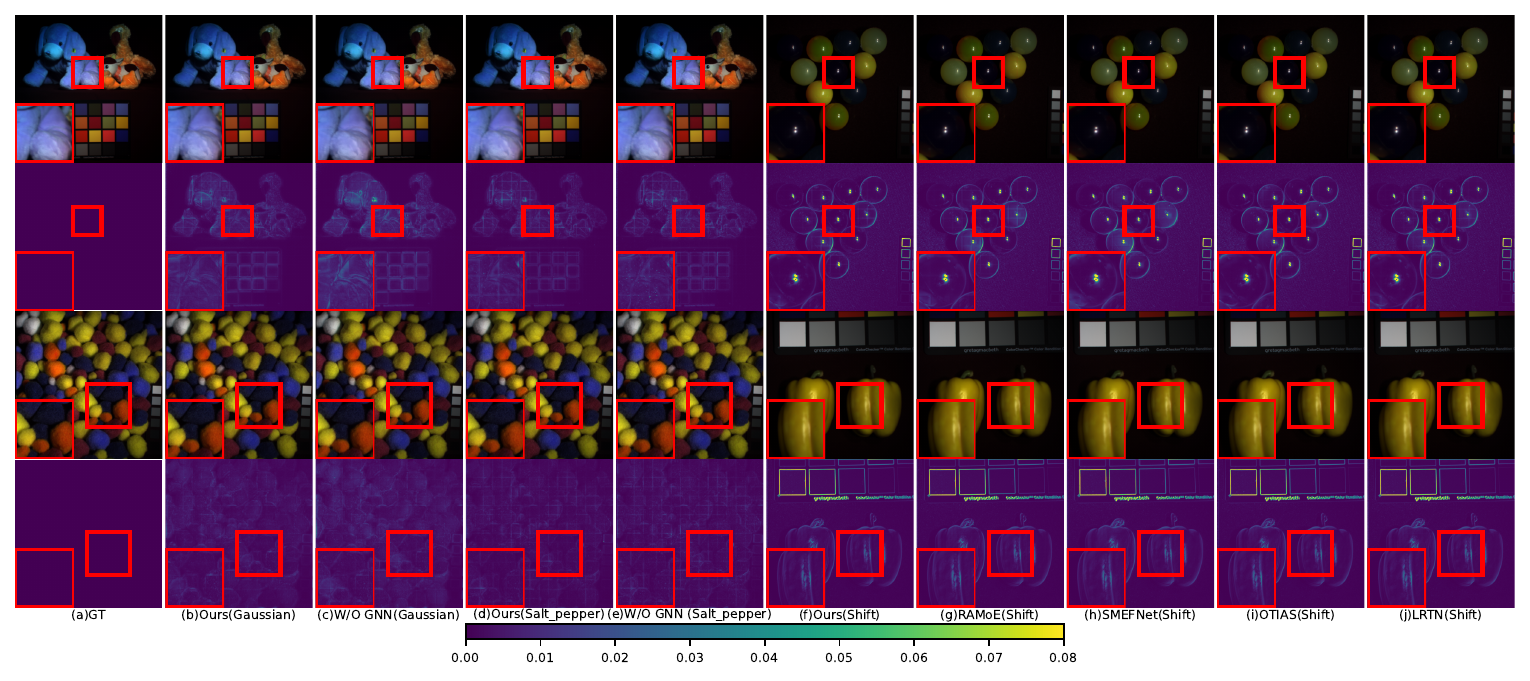}
\caption{Quantitative comparison under different MSI degradation conditions on the CAVE dataset.}
\end{figure*}

\subsection{Graph Operator Analysis}
To evaluate the contribution of different graph operators, the graph propagation module in TPFormer is replaced with different graph operator variants while all other network components remain unchanged. Specifically, GCN, GAT, and SuperGAT are individually employed for topology-aware neighborhood propagation. The quantitative comparison is summarized in Table~\ref{tab:Graph_cave}. All graph-based variants outperform the counterpart without graph propagation, demonstrating the effectiveness of topology-aware neighborhood aggregation. ``Transformer+GCN+GAT'' consistently achieves strong performance, suggesting that adaptive neighborhood weighting facilitates the learning of more discriminative feature representations. Nevertheless, the complete TPFormer consistently delivers the best performance across all evaluation metrics. The performance improvement over the pure Transformer indicates that explicit spatial topology modeling provides complementary geometric priors beyond attention-based similarity learning. While self-attention adaptively aggregates globally correlated features, topology-aware neighborhood propagation introduces complementary structural priors for local feature aggregation. Therefore, the combination of attention and graph propagation achieves a better balance between global semantic interaction and local topology preservation.

\subsection{Visualization of Learned Feature Geometry}
To further investigate the effectiveness of TPFormer in capturing spatial topology and feature manifold relationships, t-SNE visualization is performed on the intermediate embeddings extracted from TPFormer, as shown in Fig.~6. The input features exhibit scattered distributions with weak local compactness, indicating limited discriminative capability of the original spatial–spectral representations. After topology-aware neighborhood propagation, the feature embeddings become more compact and structured, demonstrating that topology-aware propagation effectively exploits local topology to refine the feature distribution. Furthermore, the output embeddings show improved compactness with reduced intra-cluster dispersion, suggesting that TPFormer enhances spatial–spectral representations by capturing intrinsic spatial–spectral structures through topology-aware aggregation.

\subsection{Cross-Dataset Generalization Evaluation}
To evaluate the generalization capability of the proposed method, cross-dataset experiments were conducted by training on the Harvard dataset and testing on the CAVE dataset. As reported in Table \ref{tab:comparison}, the proposed method maintains favorable reconstruction accuracy across multiple evaluation metrics under the cross-dataset setting. These results demonstrate robust spatial–spectral reconstruction performance under significant domain shifts, indicating strong cross-dataset generalization capability.
\begin{figure*}[ht]    
\centering     
\includegraphics[width=0.9\textwidth]{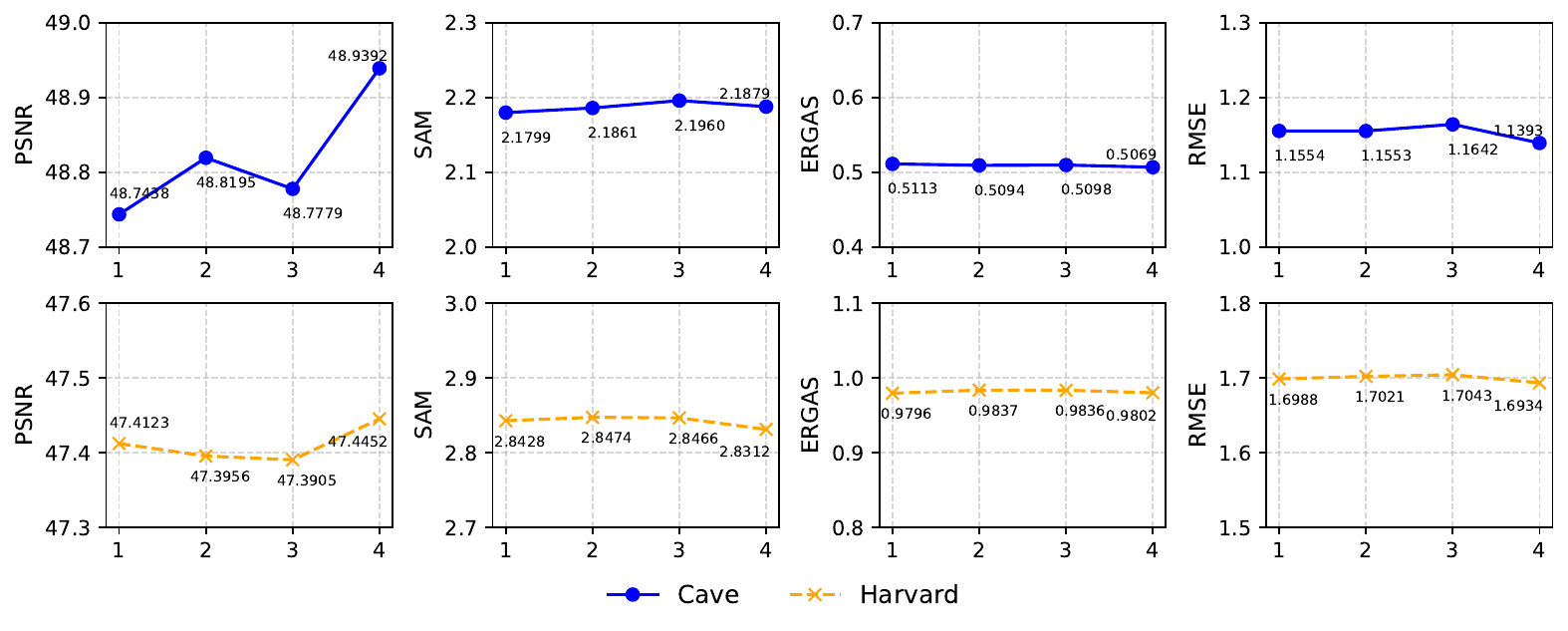}
\caption{The horizontal axis \( K \) denotes model performance with Chebyshev polynomials of varying orders.}
\end{figure*}
\begin{figure*}[htbp]    
\centering     
\includegraphics[width=0.9\textwidth]{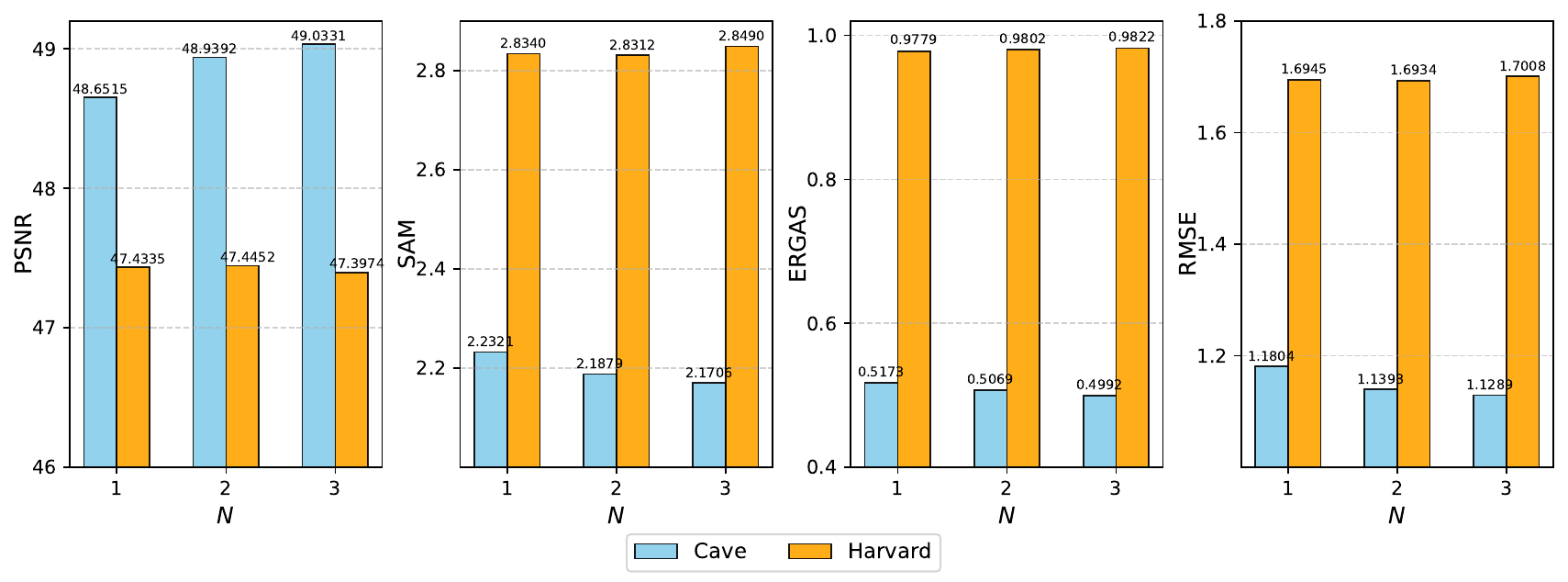}
\caption{ Impact of FDSCF block depth on fusion performance over the Harvard and CAVE datasets.}
\end{figure*}
\begin{table}[!ht]
\centering
\caption{Robustness evaluation on the Harvard-CAVE dataset.}
\label{tab:comparison}
\scriptsize
\setlength{\tabcolsep}{7pt} 
\renewcommand{\arraystretch}{1.1}
\begin{tabular}{l c c c c c}
\toprule
\textbf{Model} & \textbf{PSNR}  & \textbf{SAM}  & \textbf{ERGAS}  & \textbf{RMSE}  & \textbf{UIQI} \\
\midrule
SMGUNet \cite{YAN2025111277} & 33.1687 & 14.0351 & 3.2183 & 7.1384 & 0.8824 \\
CSGAV \cite{11014272}    & 32.1246 & 12.4105 & 3.2150 & 8.5431 & 0.9159 \\
LRTN \cite{liu2025low}     & 34.7406 & 12.0323 & 2.6809 & 6.1410 & 0.9083 \\
OTIAS \cite{deng2025otias}    & 36.2658 & 9.5623  & 2.3229 & 5.4235 & 0.9498 \\
SEMFNet \cite{11347025}     & 36.9752 &  \textbf{7.1281} & 2.0240 & 4.6617 & 0.9596 \\
RAMoE \cite{Xiao2026RAMoE}     &  \underline{38.2042} &  8.5876 &  \underline{1.8251} & 4.4126 & \textbf{0.9606} \\
Ours                          & \textbf{38.2459} & \underline{8.0694} & \textbf{1.8240} & \textbf{4.3827} & \underline{0.9597} \\
\bottomrule
\end{tabular}
\end{table}

\begin{figure*}[htbp]   
\centering     
\includegraphics[width=0.9\textwidth]{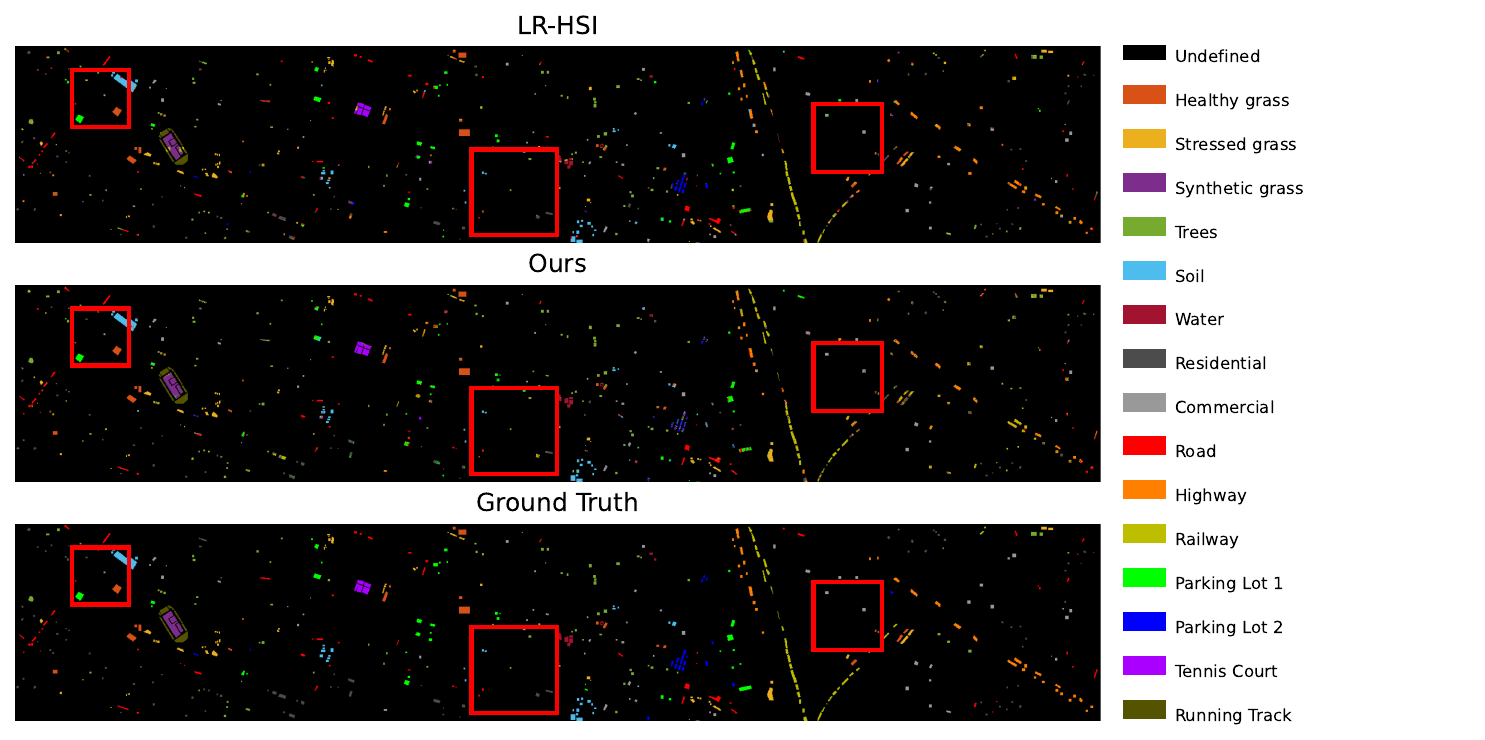}
\caption{Visual comparison of classification results on the Houston dataset.}
\end{figure*}
\begin{table}[htbp]
\centering
\caption{Quantitative robustness evaluation under different MSI degradation conditions on the CAVE dataset.}
\label{tab:results}
\resizebox{\linewidth}{!}{
\begin{tabular}{llccccc}
\toprule
\textbf{Model} & \textbf{Noise (MSI)} & \textbf{PSNR} & \textbf{RMSE} & \textbf{SAM} & \textbf{ERGAS} & \textbf{UIQI} \\
\midrule
Ours & Gaussian & \textbf{40.3873} &  \textbf{2.7043} & \textbf{5.3271} & \textbf{1.1899} & \textbf{0.9556} \\
w/o GNN & Gaussian & 39.9528 & 2.8152 & 5.3535 & 1.2179 & 0.9480 \\
Ours & Salt-pepper & \textbf{47.1649} & \textbf{1.3599} & \textbf{2.5451} & \textbf{ 0.6031} & \textbf{0.9920} \\
w/o GNN & Salt-pepper & 45.8769 & 1.5721 & 2.8295 & 0.7006 & 0.9899 \\
Ours & Shift & \textbf{36.1058} &  \textbf{4.5081} & \textbf{3.2884} & \textbf{1.9875} & \textbf{0.9600} \\
RAMoE \cite{Xiao2026RAMoE} & Shift & 36.0114 &  4.5481 & 3.3711 & 2.0024 & 0.9599 \\
SMEFNet \cite{11347025} & Shift & 35.7524 &  4.6169 & 3.7727 & 2.0325 & 0.9581 \\
OTIAS \cite{deng2025otias} & Shift & 35.9495 &  4.5627 & 3.4514 & 2.0061 & 0.9590 \\
LRTN \cite{liu2025low} & Shift & 35.7261 &  4.6980 & 3.4321 & 2.0695 & 0.9568 \\
\bottomrule
\end{tabular}
}
\end{table}

\subsection{Robustness to MSI Degradations}
To evaluate robustness against MSI degradations, Gaussian noise with a standard deviation of 0.03, salt-and-pepper noise with a density of 0.01, and a 0.5-pixel spatial shift are introduced into the MSI observations. Table~\ref{tab:results} reports the quantitative results. Under Gaussian and salt-and-pepper noise, the full model is compared with its counterpart without GNN components, where GCN, GAT, and SuperGAT are removed while all other modules remain unchanged. Under spatial misregistration, the proposed method is further compared with several state-of-the-art fusion methods.
The proposed method consistently maintains robust performance under different degradation settings. Its superiority under Gaussian and salt-and-pepper noise highlights the effectiveness of graph-based feature modeling in mitigating stochastic perturbations. Under a 0.5-pixel spatial shift, the proposed method also outperforms competing approaches, demonstrating enhanced tolerance to spatial misregistration through improved spatial–spectral correspondence. The qualitative comparisons in Fig.~7 further support these quantitative findings.
\subsection{Hyperparameter Sensitivity Analysis}
\textbf{1) The degree of the Chebyshev polynomial:} To alleviate over-smoothing in graph representation learning, Chebyshev polynomials are employed to construct a high-order graph diffusion operator, enabling a larger receptive field with stable propagation. Fig. 8 compares different polynomial orders. The fourth-order approximation achieves the best performance, as lower orders are insufficient to capture long-range dependencies, whereas higher orders introduce higher computational cost with negligible performance gains. Therefore, the fourth-order polynomial is adopted in all experiments.\par
\textbf{2) Effect of the FDSCF Block Depth:} To address the blurred structures caused by insufficient modeling of high-frequency geometric information, we propose the FDSCF module. We further evaluate the effect of stacking different numbers of FDSCF modules, as shown in Fig.~9. The results indicate that two FDSCF modules achieve the best overall performance. A single module is insufficient to capture the complex spatial–spectral interactions required for detailed structure reconstruction, whereas stacking three modules provides only marginal improvements on the CAVE dataset while slightly degrading performance on the Harvard dataset. Moreover, deeper stacking increases model complexity and computational cost with limited performance gains. Therefore, two FDSCF modules are adopted in the final architecture.

\begin{table}[htbp]
\centering
\caption{Classification Performance (\%) on the Houston Dataset with LR-HSI and Reconstructed HR-HSI.}
\label{tab:classification_results}

\renewcommand{\arraystretch}{1.0}
\setlength{\tabcolsep}{6pt}

\begin{tabular}{lcc}
\toprule

\multirow{2}{*}{\textbf{Category}} 
& \multicolumn{2}{c}{\textbf{Accuracy (\%)}} \\

\cmidrule(lr){2-3}

& \textbf{LR-HSI} 
& \textbf{Reconstructed HR-HSI} \\

\midrule

Healthy grass    & 72.4 & \textbf{81.0} \\
Stressed grass   & 69.7 & \textbf{77.2} \\
Synthetic grass  & 68.8 & \textbf{84.8} \\
Trees            & 64.2 & \textbf{80.6} \\
Soil             & \textbf{85.0} & 82.6 \\
Water            & 38.5 & \textbf{82.7} \\
Residential      & 43.4 & \textbf{45.8} \\
Commercial       & 59.3 & \textbf{61.0} \\
Road             & 60.3 & \textbf{72.2} \\
Highway          & \textbf{61.2} & 42.6 \\
Railway          & 58.7 & \textbf{68.1} \\
Parking Lot 1    & \textbf{61.5} & 57.2 \\
Parking Lot 2    & \textbf{38.1} & 22.4 \\
Tennis Court     & \textbf{87.7} & 86.3 \\
Running Track    & 82.2 & \textbf{97.5} \\

\midrule

\textbf{Average Accuracy (AA)} 
& 62.69 & \textbf{69.07} \\

\textbf{Overall Accuracy (OA)} 
& 63.62 & \textbf{68.26} \\

\bottomrule
\end{tabular}

\end{table}
\section{Performance Evaluation on Downstream Classification}
To further evaluate the practical utility of the reconstructed hyperspectral images, the proposed DDMM framework is assessed on a downstream hyperspectral classification task using the Houston dataset. Following the work of Liu \cite{liu2025low}, the LR-HSI is generated by applying Gaussian blurring with an $8\times8$ kernel ($\sigma=3$). During training, overlapping $32\times32$ image patches are extracted with a stride of 32. The network is optimized using Adam with $\beta_1=0.9$, $\beta_2=0.999$, an initial learning rate of $6\times10^{-4}$, and a cosine annealing learning-rate schedule. Training is conducted for 2000 epochs with a batch size of 4.
To quantitatively assess the discriminative quality of the reconstructed images, a support vector machine (SVM) classifier is independently trained using the interpolated LR-HSI and the reconstructed HR-HSI. Following the standard evaluation protocol, 20\% of the labeled samples are randomly selected for training, while the remaining samples are used for testing. The classification results are reported in Table~\ref{tab:classification_results}.
Compared with the interpolated LR-HSI, the reconstructed HR-HSI consistently achieves higher classification accuracy across most land-cover categories. In particular, the proposed DDMM improves the overall accuracy from 63.62\% to 68.26\% and the average accuracy from 62.69\% to 69.07\%, indicating that the reconstructed images provide more discriminative spatial-spectral representations for downstream semantic recognition, as illustrated in Fig.~10.

\section{Conclusion}
This study proposes DDMM, a dual-domain manifold modeling framework for hyperspectral image fusion that jointly explores spatial topology and pixel-level feature manifold relationships with frequency-aware spatial–spectral fusion. Specifically, TPFormer combines attention-based global interaction with topology-aware neighborhood propagation to model spatial topology and pixel-level feature manifold relationships, enabling geometry-aware spatial–spectral representation learning. Meanwhile, FDSCF performs DCT-based frequency decomposition and spatial–spectral collaborative modeling to mitigate low-frequency bias and selectively enhance high-frequency structural details, thereby improving spatial fidelity and spectral reconstruction accuracy. Extensive experiments on multiple benchmark datasets demonstrate that DDMM achieves superior overall performance over existing state-of-the-art methods in both quantitative and qualitative evaluations. Future work will investigate extending the proposed dual-domain manifold modeling paradigm to more challenging remote sensing tasks, such as cross-modal change detection and multimodal image understanding, while developing more efficient architectures for practical large-scale deployment.

{
\bibliographystyle{IEEEtran}
\bibliography{referpaper}
}

\end{document}